\documentclass[10pt,twocolumn,letterpaper]{article}

\usepackage[pagenumbers]{cvpr} 

\usepackage{marvosym}

\definecolor{cvprblue}{rgb}{0.21,0.49,0.74}
\usepackage[pagebackref,breaklinks,colorlinks,allcolors=cvprblue]{hyperref}
\usepackage[accsupp]{axessibility}

\def\paperID{*****} 
\def\confName{CVPR}
\def\confYear{2026}
\def\methodName{CrowdGaussian}
\def\diffName{CrowdRefiner}

\DeclareMathOperator{\Render}{Render}
\DeclareMathOperator{\LORM}{LORM}
\DeclareMathOperator{\SSIM}{SSIM}

\usepackage{xcolor}

\definecolor{TeaserBlue}{RGB}{65, 105, 225}
\definecolor{TeaserGreen}{RGB}{46, 139, 87}
\definecolor{TeaserYellow}{RGB}{218, 165, 32}

\title{CrowdGaussian: Reconstructing High-Fidelity 3D Gaussians\\for Human Crowd from a Single Image}

\author{
    Yizheng Song\textsuperscript{\rm 1,3} \quad
    Yiyu Zhuang\textsuperscript{\rm 1,3} \quad
    Qipeng Xu\textsuperscript{\rm 1,3} \quad
    Haixiang Wang\textsuperscript{\rm 1,3} \quad
    Jiahe Zhu\textsuperscript{\rm 1,3} \\
    Jing Tian\textsuperscript{\rm 1,3} \quad
    Siyu Zhu\textsuperscript{\rm 2} \quad
    Hao Zhu\textsuperscript{\rm 1,3,\Letter}
    \vspace{0.5em}
    \\
    \textsuperscript{\rm 1}Nanjing University, China \quad
    \textsuperscript{\rm 2}Fudan University, China
    \\
    \textsuperscript{\rm 3}State Key Laboratory of Novel Software Technology, China
    \vspace{-0.5em}
}

\begin{document}

\twocolumn[{
\renewcommand\twocolumn[1][]{#1}
\maketitle
\vspace{-25pt}
\begin{center}
    \captionsetup{type=figure}
    \includegraphics[width=\textwidth]{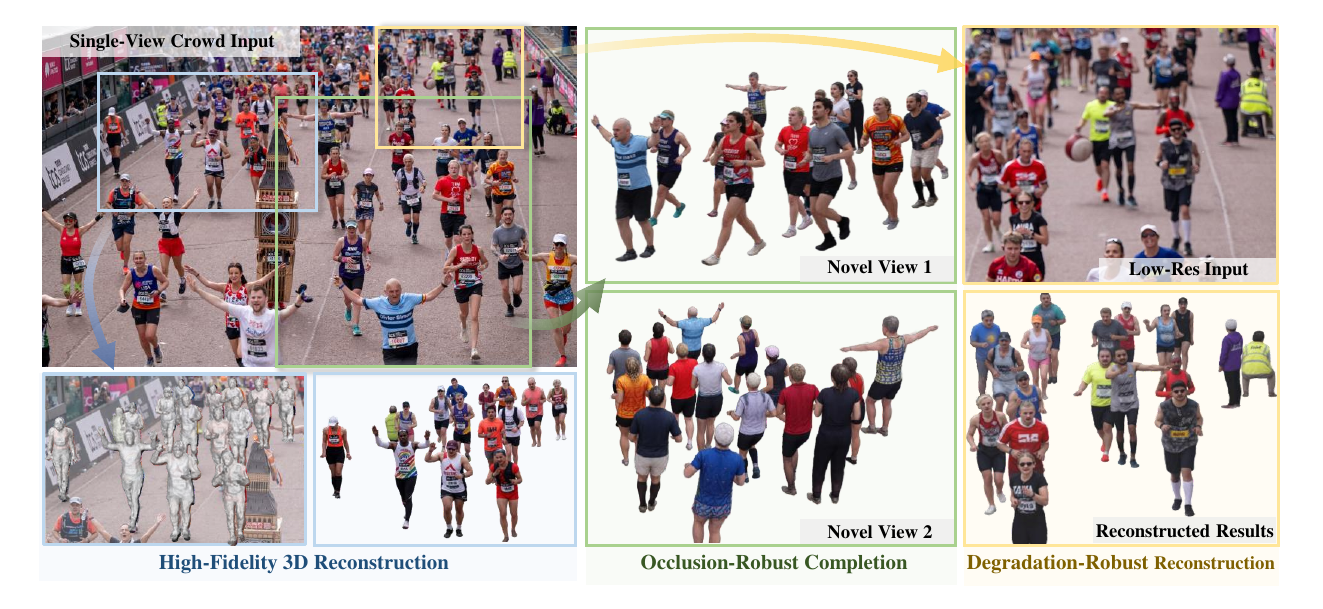}
    \captionof{figure}{
    From a single in-the-wild crowd image (\textbf{top-left}), \methodName{} reconstructs a high-fidelity multi-person 3D Gaussian scene (\textcolor{TeaserBlue}{\textbf{blue}}). 
    Crucially, our method ensures \textbf{occlusion-robust completion}, producing plausible geometry for invisible regions as verified by consistent novel-view rendering (\textcolor{TeaserGreen}{\textbf{green}}). Furthermore, it demonstrates exceptional \textbf{degradation robustness}, successfully recovering sharp details even from low-resolution inputs (\textcolor{TeaserYellow}{\textbf{yellow}}).
    }
    \label{fig:teaser}
\end{center}
}]

\maketitle
\begin{abstract}

Single-view 3D human reconstruction has garnered significant attention in recent years. Despite numerous advancements, prior research has concentrated on reconstructing 3D models from \textbf{clear}, \textbf{close-up} images of \textbf{individual subjects}, often yielding subpar results in the more prevalent multi-person scenarios. Reconstructing 3D human crowd models is a highly intricate task, laden with challenges such as: 1) extensive occlusions, 2) low clarity, and 3) numerous and various appearances. To address this task, we propose \methodName, a unified framework that directly reconstructs multi-person 3D Gaussian Splatting (3DGS) representations from single-image inputs. To handle occlusions, we devise a self-supervised adaptation pipeline that enables the pretrained large human model to reconstruct complete 3D humans with plausible geometry and appearance from heavily occluded inputs.
Furthermore, we introduce Self-Calibrated Learning (SCL). This training strategy enables single-step diffusion models to adaptively refine coarse renderings to optimal quality by blending identity-preserving samples with clean/corrupted image pairs. The outputs can be distilled back to enhance the quality of multi-person 3DGS representations. Extensive experiments demonstrate that \methodName{} generates photorealistic, geometrically coherent reconstructions of multi-person scenes.

\end{abstract}    
\section{Introduction}
\label{sec:intro}

Creating high-fidelity, animatable 3D human avatars from monocular images is fundamental to advancing immersive applications in virtual reality, teleconferencing, and digital content creation. Recent breakthroughs in single-image 3D human reconstruction have demonstrated remarkable progress~\cite{zhang2024sifu, li2025pshuman, zhuang2025idol, qiu2025lhm, huang2025adahuman, chen2025synchuman}. However, reconstructing the 3D shape and appearance of human crowds from a single image remains an unexplored territory, a scenario that is prevalent but overlooked.

Performing full-body 3D modeling for each individual in a crowd from a single crowd image poses a non-trivial challenge.
An intuitive strategy is to integrate human pose estimation algorithms with 3D reconstruction techniques. Specifically, this approach first leverages human pose estimation algorithms to accurately extract crowd pose information, a task well within current capabilities~\cite{sun2022putting, sun2023trace, wen2023crowd3d, sun2024aios, huang2024rcr, baradel2024multi, wang2025prompthmr}. Following this, it crops individual images of each person and leverages single-image human 3D reconstruction algorithms to independently model each person, culminating in the comprehensive 3D reconstruction of the entire crowd. 

However, this strategy faces three fundamental challenges. Firstly, in densely populated human crowds, frequent and severe occlusions occur both between individuals and between individuals and objects, often resulting in irregularly absent or obscured body parts. Due to varying person scales and complex occlusion patterns, applying 2D inpainting~\cite{zhou2021human, liang2024dmat, liu2024object} to recover complete appearances is highly challenging and often produces semantically implausible or geometrically inconsistent completions. Directly feeding such incomplete inputs into existing single-image 3D reconstruction methods yields incomplete geometries with visible transparency artifacts from unobserved regions. Secondly, due to limited 3D supervision and the lack of pose and texture diversity in training datasets, large human models struggle to recover high-frequency details when processing low-resolution person crops, leading to a blurred appearance and a lack of fine details.  Lastly, this task places higher demands on efficiency due to the large number of humans to be reconstructed at a single attempt. A unified solution is preferred over concatenating individually trained models for efficient and effective reconstruction.

To address these challenges, we propose \methodName, a unified two-stage framework for reconstructing the 3D shape and appearance of multiple humans from a single image. In the first stage, we estimate per-person 3D poses and locations using multi-person HMR~\cite{wang2025prompthmr}, then segment each individual via SAM~\cite{kirillov2023segment}. From these (potentially occluded) person crops, we generate an initial coarse multi-person 3DGS~\cite{kerbl20233d} representation using LORM (Large Occluded Human Reconstruction Model), which is derived from a pre-trained full body large human model~\cite{qiu2025lhm} through a self-supervised adaptation framework: by constructing paired data—clean renderings from the frozen teacher and occluded inputs—we distill robust completion capability into the student model without requiring 3D annotations. This enables LORM to recover complete geometry and texture even under severe occlusions.

While LORM produces geometrically complete reconstructions, the results may lack high-frequency detail due to low-resolution inputs. Inspired by diffusion-based 3DGS refinement methods~\cite{wu2025difix3d+, wu2025genfusion, wei2025gsfix3d, yin2025gsfixer}, we introduce \diffName{}, a single-step diffusion model that refines coarse crowd 3DGS representation by leveraging 2D generative priors. We train \diffName{} with the Self-Calibrated Learning (SCL) strategy, a novel approach that combines identity-preserving samples with degradation-to-clean pairs. This enables adaptive enhancement—refining under-recovered regions while preserving well-structured areas—without requiring explicit segmentation masks or external labels. The refined outputs serve as pseudo-ground truths for differentiable rendering optimization, resulting in sharper textures and improved local coherence across the multi-person scene.

In summary, our main contributions are:
\begin{itemize}[itemsep=0pt,parsep=0pt,topsep=2bp]
    \item We present a robust framework for single-image 3D human reconstruction tailored for \textbf{crowd scenes}—a common yet under-explored scenario. 
    Our method achieves significantly superior reconstruction quality compared to state-of-the-art methods, particularly in complex multi-person arrangements.
    \item We propose a self-supervised fine-tuning strategy that empowers large human reconstruction models to restore complete 3D geometries from occluded inputs, eliminating the need for additional 3D annotations.
    \item We introduce \diffName{}, a single-step diffusion-based refiner trained via Self-Calibrated Learning (SCL). This strategy adaptively enhances 2D renderings from reconstructed 3D Gaussians by balancing refinement strength to prevent over-correction, yielding pseudo-ground truth for 3D quality improvement.
\end{itemize}
\section{Related Work}
\label{sec:relatedwork}

\begin{figure*}[tb] \centering
    \includegraphics[width=\textwidth]{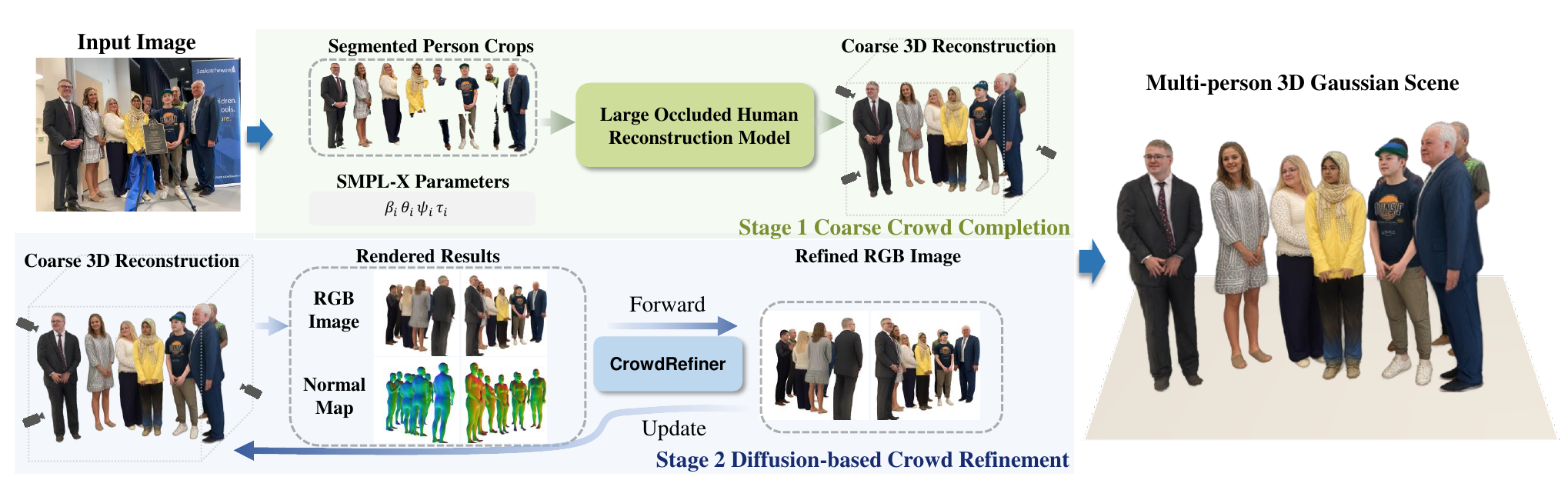}
    \vspace{-20pt}
    \caption{
    Overview of the proposed \methodName{} framework. Our pipeline operates in two stages. In \textcolor{TeaserGreen}{\textbf{Stage 1}}, we first estimate SMPL-X parameters and segment individuals from the input image. These occluded crops are processed by our \textbf{LORM} to hallucinate complete geometries, assembling an initial coarse multi-person 3DGS scene. In \textcolor{TeaserBlue}{\textbf{Stage 2}}, we render this coarse scene into RGB images and normal maps. Our \diffName{} leverages these cues to generate high-fidelity pseudo-ground truths, which are then distilled back into the 3D Gaussians via differentiable rendering, significantly enhancing local details and overall sharpness.
    } \label{fig:pipeline}
    \vspace{-15pt}
\end{figure*}

\subsection{Single-Image Human Reconstruction}

Early monocular 3D human reconstruction methods primarily relied on parametric models such as SMPL~\cite{loper2023smpl, pavlakos2019expressive} to predict vertex offsets for clothed humans~\cite{alldieck2018detailed, zhang2021pymaf, pavlakos2019texturepose}, but are constrained by fixed topology and struggle with complex garments or hairstyles. Subsequent works leveraged implicit representation methods~\cite{saito2019pifu, saito2020pifuhd, huang2020arch, he2021arch++, zheng2021pamir, zhang2024sifu}, enabling arbitrary topologies and providing better flexibility but struggling with complex poses. Recent advances in large reconstruction models (LRMs)~\cite{hong2023lrm, tang2024lgm} have enabled fast, feed-forward 3D human reconstruction from single images~\cite{weng2024template, zhang20243gen, li2025pshuman, huang2025adahuman, qiu2025lhm, zhuang2025idol, chen2025synchuman}. Leveraging transformer architectures and large-scale datasets, these models reduce inductive bias and improve generalization to diverse poses and clothing.

However, existing methods often produce incomplete geometric and textural structures when processing irregularly occluded inputs—a common scenario in multi-person scenes. Some approaches attempt to reconstruct 3D humans from incomplete images~\cite{wang2023complete, dutta2025chrome}. Wang $\mathit{et\ al.}$~\cite{wang2023complete} proposes a coarse-to-fine geometric recovery pipeline followed by progressive texture inpainting to synthesize missing appearances. This multi-stage approach is computationally expensive in a multi-person reconstruction scenario. Moreover, its reliance on limited 3D supervision weakens generalization to diverse poses and clothing. CHROME~\cite{dutta2025chrome} employs multi-view diffusion to synthesize occlusion-free images under pose control, followed by a 3D reconstruction module that predicts Gaussian representations from the generated views. However, due to subtle misalignments and inconsistencies across the synthesized multi-view images, the resulting 3D scene exhibits texture corruption—particularly around facial regions where fine details are critical. 

To address these issues, we introduce \textbf{LORM} (Large Occluded Human Reconstruction Model). Adapted from pre-trained large human reconstruction models~\cite{qiu2025lhm} via a self-supervised framework, LORM enables complete and robust 3D human recovery even from heavily occluded inputs.

\subsection{3D Gaussian Refinement}
Very recently, 3D Gaussian Splatting~\cite{kerbl20233d} has been widely used for 3D human body representation~\cite{zhuang2025idol, zhang20243gen, qiu2025lhm, zhu2026gshvv}. Several works explore incorporating geometric priors such as depth to improve the geometric fidelity of 3D Gaussians~\cite{zhu2024fsgs, kumar2024few}. Recent approaches further leverage 2D diffusion models as priors to enhance 3D reconstruction quality. Early approaches use diffusion models as iterative optimizers, with denoising steps guiding the update of 3D Gaussians~\cite{warburg2023nerfbusters, wu2024reconfusion, xie2023sparsefusion}, though this incurs a high computational cost due to repeated sampling. To improve efficiency, several methods adopt a pseudo-observation paradigm, refining coarse renderings into high-fidelity images for 3D optimization supervision~\cite{liu2024deceptive, liu20243dgs, wu2025genfusion, wu2025difix3d+, wei2025gsfix3d, yin2025gsfixer, prinzler2025joker}.

Deceptive-3DGS~\cite{liu2024deceptive} refines sparse-view renderings using a diffusion model guided by uncertainty estimates to enhance multi-view consistency. 3DGS-Enhancer~\cite{liu20243dgs} trains a video diffusion model on paired low- and high-quality image sequences for temporal coherence in Gaussian Splatting. GenFusion~\cite{wu2025genfusion} fine-tunes a video diffusion model with masked inpainting to repair view-dependent artifacts in RGB-D sequences. DIFIX~\cite{wu2025difix3d+} employs a single-step diffusion model trained on synthetically corrupted image pairs to correct rendering artifacts. GSFix3D~\cite{wei2025gsfix3d} adapts a pretrained diffusion model via task-specific fine-tuning for joint artifact removal and inpainting. GSFIXER~\cite{yin2025gsfixer} introduces a reference-guided video diffusion model that leverages multi-frame cues and geometric features to restore degraded 3DGS renders. Joker~\cite{prinzler2025joker} uses a 2D diffusion prior to optimize 3D heads with extreme expressions, enabling view-consistent synthesis of features like tongue articulation.

Unlike general-scene methods, we focus on enhancing human crowd 3D Gaussians, where identity preservation and facial details are paramount.
We introduce \diffName{}, a single-step refiner trained via Self-Calibrated Learning to adaptively balance structural preservation and detail enhancement.
The resulting high-fidelity pseudo-ground truths are then distilled back into the 3D Gaussians, significantly improving geometric and textural fidelity.

\section{Method}
\label{sec:method}

\begin{figure}[tb] \centering
    \includegraphics[width=0.48\textwidth]{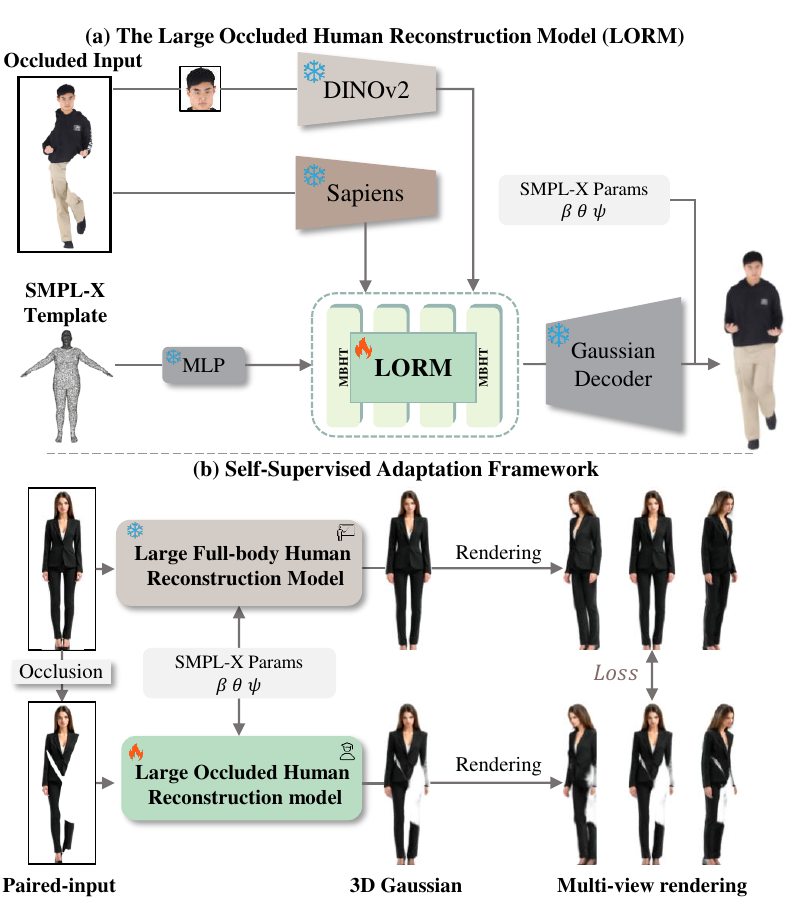}
    \vspace{-15pt}
    \caption{
The Large Occluded Human Reconstruction Model (LORM). (a) Architecture. LORM takes an occluded image and a template to reconstruct a complete 3D human. To preserve priors while enabling efficient adaptation, we freeze pre-trained backbones and inject trainable LoRA exclusively into the transformer. (b) Self-Supervised Training. We employ a Teacher-Student framework where the teacher generates clean pseudo-GTs from complete images. These signals guide the student to hallucinate complete geometries from occluded inputs via self-distillation, achieving robustness without external 3D supervision.
    } 
    \vspace{-15pt}
    \label{fig:LORM_training}
\end{figure}

\subsection{Overview}

Given an image $ I \in \mathbb{R}^{H \times W \times 3} $ containing $N$ people, our goal is to reconstruct the 3D shape and appearance of all individuals into coherent 3D Gaussian Splatting (3DGS) representations. As illustrated in Figure~\ref{fig:pipeline}, we propose \methodName, a unified framework for single-image multi-person 3D reconstruction.

Our pipeline proceeds in two main stages: \emph{Coarse Crowd Completion} and \emph{Diffusion-based Crowd Refinement}.

In \emph{Coarse Crowd Completion}, we estimate per-person 3D poses and camera-space positions using a multi-person HMR model~\cite{wang2025prompthmr}. Guided by these estimates, we segment each individual from the background using SAM~\cite{kirillov2023segment}, yielding $N$ person crops that are often incomplete due to occlusions. 
We then generate an initial coarse multi-person 3DGS scene by applying the Large Occluded Human Reconstruction Model (LORM) to each segmented crop—enabling complete geometry recovery despite missing regions. This stage is detailed in Section~\ref{sec:occlusion_free_recon}.

In \emph{Diffusion-based Crowd Refinement}, we enhance the reconstructed multi-person 3D Gaussians using a single-step diffusion model. To prevent over-refinement, we train it with a Self-Calibrated Learning (SCL) strategy that adaptively preserves well-recovered regions while enhancing under-refined details. The refined outputs serve as high-fidelity pseudo-ground truths, which are distilled back into the 3D Gaussians via differentiable rendering to improve both local detail and global coherence. This refinement pipeline is described in Section~\ref{sec:diffusion_refine}.

\begin{figure}[tb] \centering
    \includegraphics[width=0.48\textwidth]{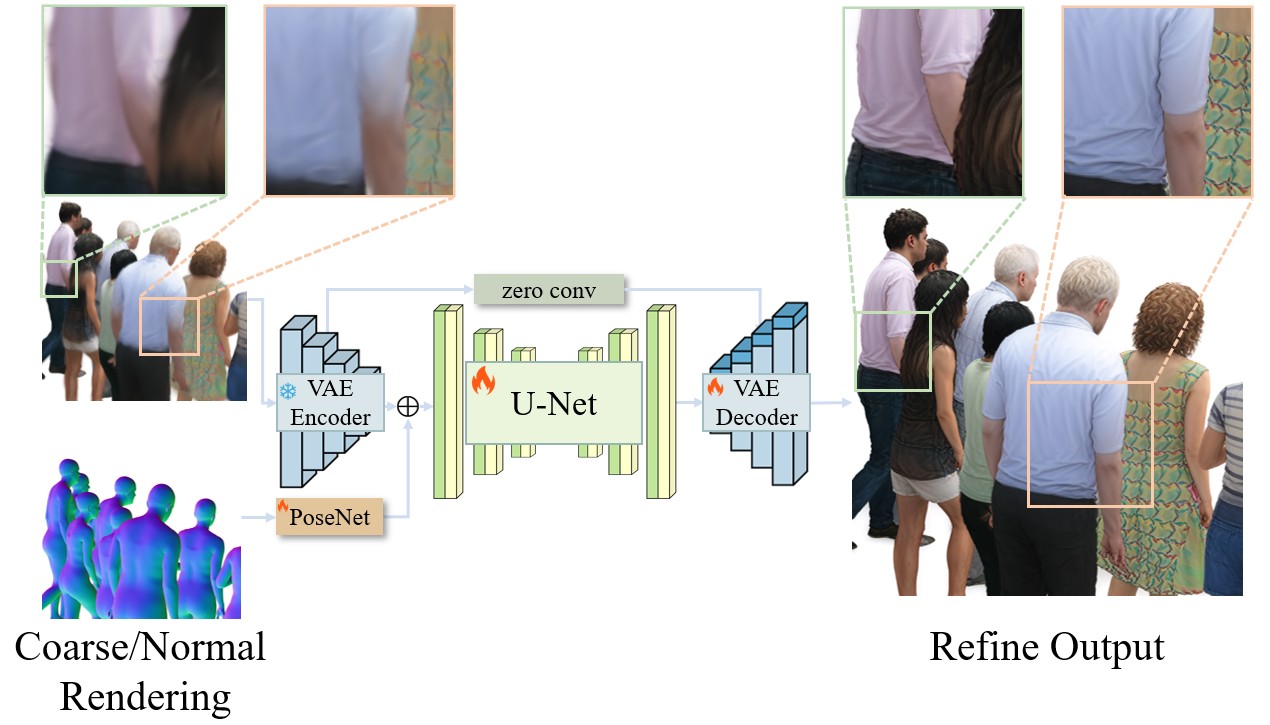}
    \vspace{-15pt}
    \caption{Architecture of \diffName{}, our single-step diffusion refiner for Crowd 3D Gaussians enhancement. 
    Given a coarse rendering  and its corresponding SMPL normal map as geometric prior, \diffName{} generates a high-fidelity refined output. 
    The model is fine-tuned from SD-Turbo with a LoRA-finetuned VAE decoder and a trainable PoseNet, while the VAE encoder remains frozen. 
    Zoom-in regions highlight significant improvements in local details such as hair and clothing textures.} 
    \vspace{-15pt}
    \label{fig:fig_diff}
\end{figure}

\subsection{Occlusion-free 3D Human Reconstruction}
\label{sec:occlusion_free_recon}

\noindent \textbf{Preliminaries.}
We build upon a pre-trained large human reconstruction model~\cite{qiu2025lhm} as our foundation, which takes a complete human image as input and outputs a 3D human represented as a set of 3D Gaussian:
\begin{equation}
\mathcal{G} = \left\{ (\boldsymbol{\mu}_i, \boldsymbol{\Sigma}_i, \alpha_i, \mathbf{c}_i) \right\}_{i=1}^N,
\end{equation}
where $\boldsymbol{\mu}_i \in \mathbb{R}^3$ is the center, $\boldsymbol{\Sigma}_i \in \mathbb{R}^{3\times3}$ the covariance, $\alpha_i \in [0,1]$ the opacity, and $\mathbf{c}_i \in \mathbb{R}^3$ the view-dependent color. The model employs a Multimodal Body-Head Transformer (MBHT) to fuse SMPL-based geometric priors with visual features extracted by the \textit{frozen} Sapiens encoder.

Though the Sapiens encoder—being a Masked Autoencoder (MAE)~\cite{he2022masked}—possesses inherent potential for handling partial visibility, the human reconstruction model lacks occlusion-aware training. Consequently, it fails to produce coherent geometries when processing occluded inputs, as its transformer backbone struggles to integrate incomplete visual features. To address this limitation, we propose LORM, a \textbf{L}arge \textbf{O}ccluded Human \textbf{R}econstruction \textbf{M}odel, which is obtained by applying our parameter-efficient self-supervised adaptation framework to the pre-trained model.

\noindent \textbf{Occlusion-free 3D Human Reconstruction.}
We propose the Large Occluded Human Reconstruction Model (LORM) to recover complete 3D geometries from occluded single-view inputs ($I_{\text{occ}}$), as shown in Figure~\ref{fig:LORM_training} (a). To leverage robust 2D-to-3D generative priors, we initialize LORM with a pre-trained large reconstruction model~\cite{qiu2025lhm}. However, this baseline fails on occlusions due to domain gaps. Fine-tuning with external 3D supervision is suboptimal, as it often amplifies geometric bias inherent in monocular ambiguity, damaging pre-trained priors. To circumvent this, we introduce a self-supervised teacher-student framework that adapts the model using only single-view images, ensuring geometric consistency without external 3D annotations.

\begin{figure}[tb] \centering
    \includegraphics[width=0.48\textwidth]{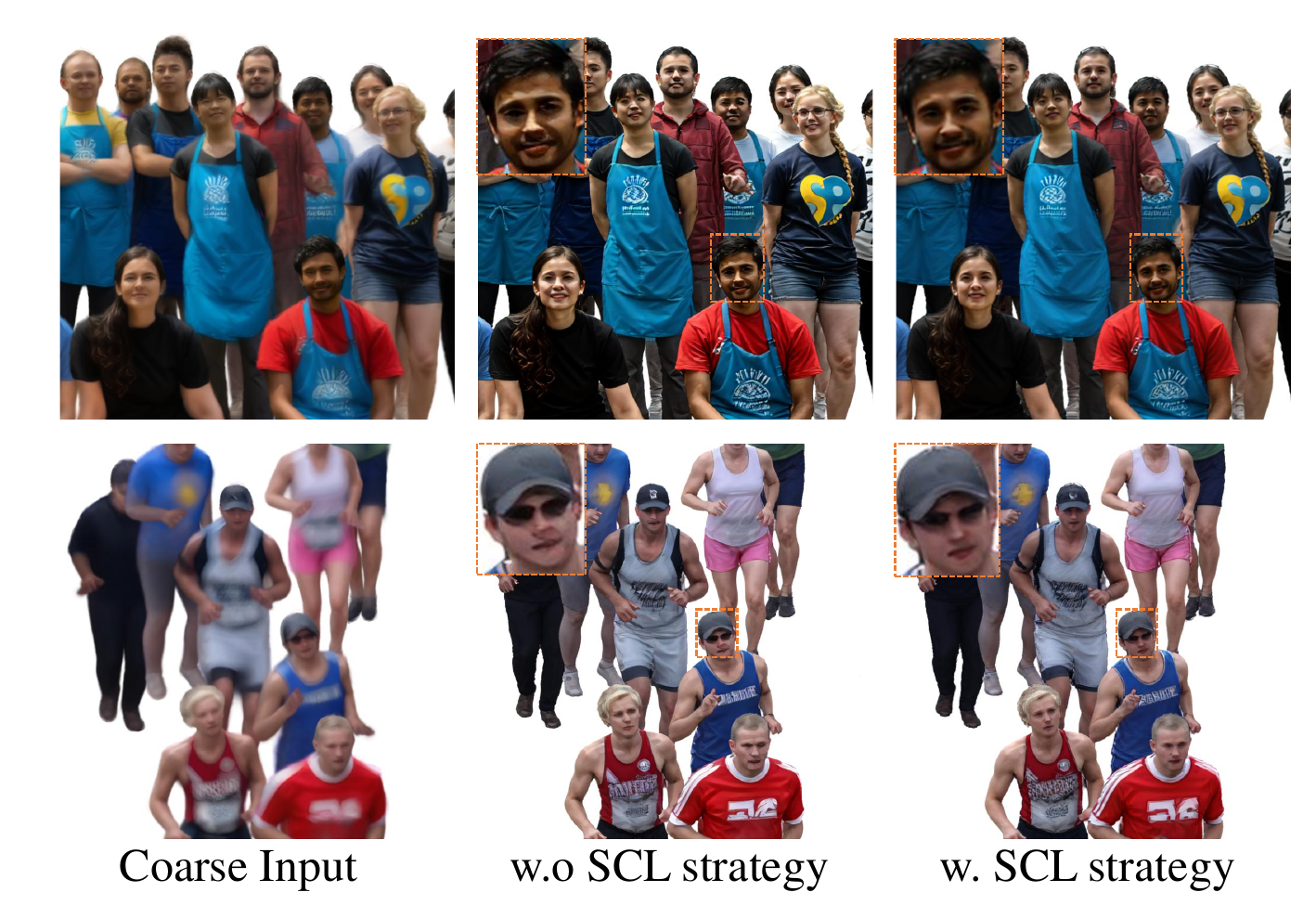}
    \vspace{-25pt}
    \caption{Effect of Self-Calibrated Learning (SCL). Without SCL (middle), the model tends to over-refine, causing facial distortions and artifacts. With SCL (right), structural integrity is preserved while details are enhanced, demonstrating adaptive refinement enabled by mixed identity supervision during training.} 
    \vspace{-20pt}
    \label{fig:SCL_vis}
\end{figure}

\noindent \textbf{Self-Supervised Adaptation Framework.}
The training framework is illustrated in Figure~\ref{fig:LORM_training} (b). We construct training pairs from thousands of frontal images from the HuGe100K dataset~\cite{zhuang2025idol}.
In the teacher stream, the \textit{frozen} pre-trained model processes a complete image $I_{\text{full}}$ to yield a complete 3D Gaussian representation $\mathcal{G}_{\text{full}}$.
We render $\mathcal{G}_{\text{full}}$ from $V$ novel viewpoints to generate clean pseudo-ground truths:
\begin{equation}
R_{\text{clean}}^{(v)} = \Render(\mathcal{G}_{\text{full}}, \theta_v), \quad v = 1,\dots,V, 
\end{equation}
where $\theta_v$ denotes the camera pose for the $v$-th view.

In the student stream, to simulate realistic occlusions, we generate $I_{\text{occ}}$ by applying irregular masks—combining random Bézier curve-based erasures and keypoint-centered ellipses—to $I_{\text{full}}$. The student model (LORM) predicts 3D Gaussians from $I_{\text{occ}}$, which are rendered into coarse views:
\begin{equation}
R_{\text{coarse}}^{(v)} = \Render(\LORM(I_{\text{occ}}, \theta), \theta_v).
\end{equation}

We optimize the adapted parameters using a self-distillation loss that aligns the coarse student outputs with the clean teacher renderings:
\begin{equation}
\begin{aligned}
\mathcal{L}_{\text{self-distill}} = \sum_{v=1}^V \bigg( & \lambda_{\text{rgb}} \| R_{\text{clean}}^{(v)} - R_{\text{coarse}}^{(v)} \|_2 \\
& + \lambda_{\text{ssim}} \left(1 - \SSIM(R_{\text{clean}}^{(v)}, R_{\text{coarse}}^{(v)})\right) \bigg),
\end{aligned}
\label{eq:self-distill}
\end{equation}
where $\lambda_{\text{rgb}}$ and $\lambda_{\text{ssim}}$ are balancing weights.

To minimize disruption to pre-trained visual priors while maintaining efficiency, we inject trainable Low-Rank Adaptation (LoRA) modules~\cite{hu2022lora} exclusively into the MBHT transformer. The Sapiens encoder and Gaussian decoder remain frozen. This design preserves robust feature extraction while focusing adaptation solely on hallucinating complete 3D structures from incomplete features.

\begin{figure*}[tb] \centering
    \includegraphics[width=\textwidth]{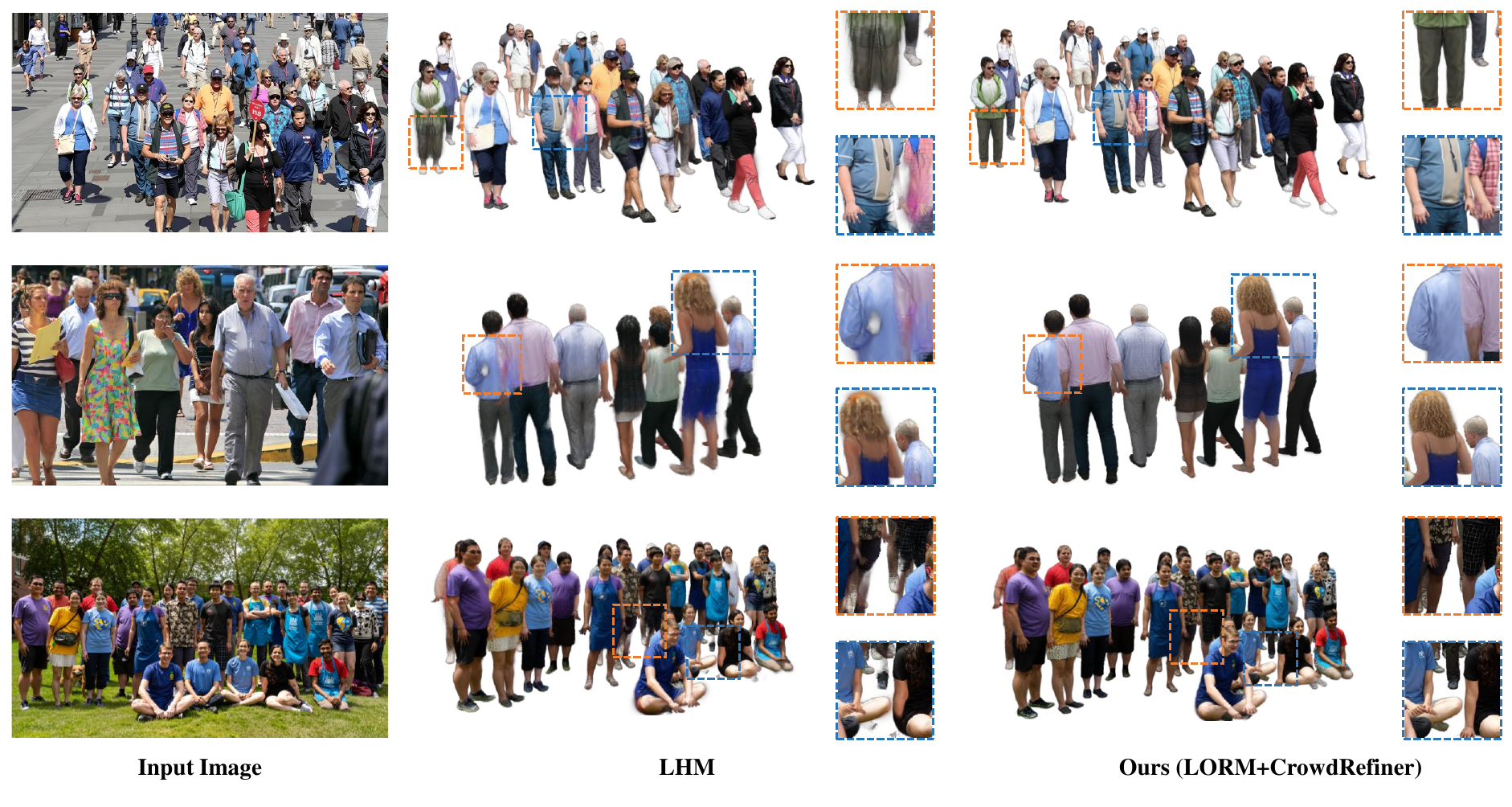}
    \vspace{-25pt}
    \caption{
    Single-shot multi-person 3D reconstruction.  While the LHM baseline (\textbf{middle}) struggles with artifacts and missing details in occluded regions, our method (\textbf{right}) generates complete geometry and high-fidelity textures. Zoom-ins highlight our superior recovery of faces and hands, validating robustness to occlusion and low-resolution inputs.
   }
    \label{fig:exp_fig1}
    \vspace{-15pt}
\end{figure*}

\subsection{Multi-Person Scene Refinement}
\label{sec:diffusion_refine}
Although LORM recovers complete geometries, the initial crowd 3DGS representation often suffers from over-smoothing and lacks high-frequency details due to the resolution limits of the reconstruction model. To address this, we introduce \diffName{}, a single-step diffusion model specifically trained to act as a high-fidelity detail enhancer. 
Unlike standard diffusion models, \diffName{} is trained to learn a geometry-aware refinement policy and is applied \textit{online} to distill details back into the crowd 3DGS via an iterative optimization pipeline.

\noindent \textbf{\diffName{} Design and Training.}
Built upon SD-Turbo~\cite{sauer2024adversarial, parmar2024one, wu2025difix3d+}, \diffName{} is designed for efficient single-step inference, as shown in Figure~\ref{fig:fig_diff}. To ensure geometric consistency, the model is addtionally conditioned on its corresponding SMPL normal map $N$ to the coarse RGB rendering $R_{\text{coarse}}$.
Specifically, we encode the normal map via a lightweight pose encoder and the RGB image through the VAE encoder. These multi-modal features are then injected into the UNet backbone to guide the generation process. To preserve fine-grained image details during refinement, we fine-tune it using LoRA adapters injected into the decoder, while keeping the VAE encoder frozen.

Training is performed on a synthetic dataset constructed from THuman2.1~\cite{zheng2019deephuman}, comprising pairs of LORM-generated coarse renderings and high-quality ground truths. However, we observe that standard supervised training on these degradation-to-clean pairs often leads to \textbf{over-refinement}, resulting in facial distortions and artifacts (as shown in Figure~\ref{fig:SCL_vis}). 

To mitigate this, we introduce the \textbf{Self-Calibrated Learning (SCL) strategy}. This approach fosters a \textit{balanced refinement} policy by randomly mixing standard degradation pairs $(R_{\text{coarse}}, R_{\text{gt}})$ with identity-preserving samples $(R_{\text{gt}}, R_{\text{gt}})$ during training. This mixed supervision scheme encourages the model to \textit{selectively} enhance under-recovered regions while strictly preserving well-structured areas.

The model is trained with a composite loss:
\begin{equation}
\begin{split}
\mathcal{L}_{\text{diff}} = & \lambda_{L2} \mathcal{L}_{\text{L2}} + \lambda_{\text{lpips}} \mathcal{L}_{\text{LPIPS}} \\
                              & + \lambda_{\text{ssim}} \mathcal{L}_{\text{SSIM}} + \lambda_{\text{gram}} \mathcal{L}_{\text{Gram}},
\end{split}
\label{eq:diff}
\end{equation}
where the L2, LPIPS, and SSIM losses ensure pixel-level accuracy and perceptual similarity. 
A Gram loss~\cite{wu2025difix3d+} is additionally applied to enforce texture sharpness by matching high-order feature statistics.

\noindent \textbf{Crowd 3D Gaussians Refinement.}
During the inference stage, we apply the pre-trained \diffName{} to enhance the target crowd scene. 
To avoid the prohibitive computational costs associated with processing individuals sequentially, we adopt a holistic strategy to refine local crowd segments simultaneously.

Specifically, to handle complex spatial arrangements, we first group individuals into spatially coherent clusters using DBSCAN~\cite{schubert2017dbscan} based on their root positions.
For each cluster, we render coarse views $R_{\text{coarse}}^{(v)}$ and process them through \diffName{} to generate high-fidelity pseudo-ground truths $R_{\text{refined}}^{(v)}$.
 
 Finally, we distill these refined priors back into the 3D Gaussians by iteratively minimizing the difference between the current rendering and the refined target:
\begin{multline}
\mathcal{L}_{\text{optim}} = \| R_{\text{refined}}^{(v)} - R_{\text{coarse}}^{(v)} \|_1 \\
+ \lambda_{\text{ssim}} (1 - \text{SSIM}(R_{\text{refined}}^{(v)}, R_{\text{coarse}}^{(v)})).
\end{multline}
This distillation process effectively transfers the generative details into the 3D representation, significantly improving geometric sharpness and local fidelity across the crowd.

\section{Experiments}
\label{sec:exp}

\begin{figure*}[tb] \centering
    \includegraphics[width=\textwidth]{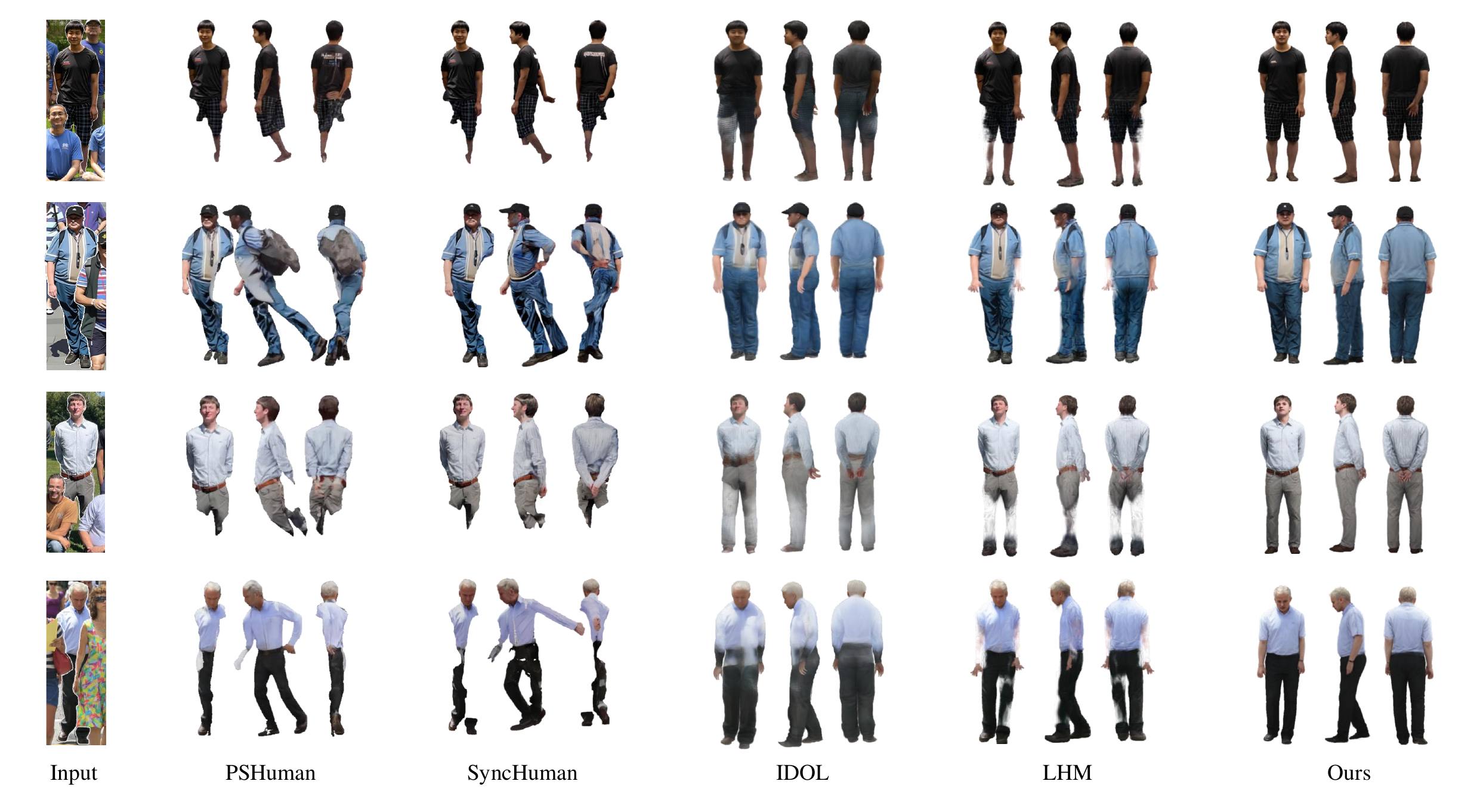}
    \vspace{-25pt}
    \caption{
    Qualitative comparison on occluded in-the-wild images. 
    Mesh-based methods (PSHuman, SyncHuman) fail to recover complete geometry. 
    3DGS-based approaches (IDOL, LHM) produce transparency artifacts or incoherent and distorted textures in missing regions.
    In contrast, our method reconstructs complete 3D geometry and texture, demonstrating robustness to real-world occlusion.
    } \label{fig:exp_fig2}
    \vspace{-15pt}
\end{figure*}

\subsection{Implementation Details}
\label{sec:impl_details}

\noindent \textbf{Data Construction for LORM Training.}
We select 1,002 frontal images from HuGe100K~\cite{zhuang2025idol} for the training of LORM. We initialize  our model using the pre-trained weights from {LHM-500M-HF}. During end-to-end optimization, LORM renders its 3DGS representation from 24 fixed camera viewpoints using differentiable rendering.

\noindent \textbf{Data Construction for \diffName{} Training.}
For the training of \diffName{}, we utilize a paired dataset constructed with identities from THuman2.1~\cite{zheng2019deephuman}. We built 114 synthetic multi-person scenes by composing the high-quality meshes in diverse spatial configurations. For each scene, we generate paired renderings from 126 viewpoints on a hemisphere: (1) the high-fidelity ground-truth $R_{\text{gt}}$ is rendered from the original THuman meshes, (2) the corresponding coarse input $R_{\text{coarse}}$ is rendered from the 3DGS representation reconstructed by our Stage 1 LORM. This process yields our final $(R_{\text{coarse}}, R_{\text{gt}})$ training pairs, (3) the normal map from SMPL-X model. 
Of these 114 scenes, 91 (80\%) are used for training and 23 for testing.

\subsection{Comparisons}
\label{sec:eval_results}

\noindent \textbf{Multi-Person Reconstruction.}
Figure~\ref{fig:exp_fig1} presents qualitative results on challenging in-the-wild crowd scenes. Despite severe inter-person and object occlusions, as well as the low resolution of individual person crops, our framework successfully reconstructs high-fidelity multi-person 3DGS representations. 
Compared to the baseline LHM—where missing regions often result in transparency artifacts and blurred details (e.g., facial features, hands)—our full pipeline leverages LORM for robust structural completion and \diffName{} for texture refinement. This synergy produces complete geometry with sharp, coherent appearance. Zoom-in views highlight significant improvements in fine-scale details, demonstrating the effectiveness of our approach in recovering both shape and appearance under real-world conditions.

\begin{table}[t]
\small
\centering
\caption{Quantitative comparison on occluded human reconstruction (THuman2.1) with randomly applied masks.}
\label{tab:lhmtune_quant}
\begin{tabular}{l c c c}
\toprule
Method & PSNR $\uparrow$ & SSIM $\uparrow$ & LPIPS $\downarrow$ \\
\midrule
IDOL~\cite{zhuang2025idol} & 18.063 & 0.919 & 0.994 \\
LHM~\cite{qiu2025lhm} & 18.171 & 0.918 & 1.012 \\
LORM & 18.566 & 0.923 & 0.956 \\
LORM+\diffName{} & \textbf{18.619} & \textbf{0.931} & \textbf{0.914} \\
\bottomrule
\vspace{-20pt}
\end{tabular}
\end{table}

\begin{table}[t]
\small
\centering

\caption{Quantitative comparison under increasing occlusion ratios (20\%--60\%) on THuman2.1.}
\label{tab:occ_ratio}
\begin{tabular}{l c ccc}
\toprule
Method & Occlusion Ratio & PSNR $\uparrow$ & SSIM $\uparrow$ & LPIPS $\downarrow$ \\
\midrule
 & 20\% & 18.196 & 0.921 & 0.978 \\
IDOL & 40\% & 17.412 & 0.914 & 1.027 \\
 & 60\% & 16.667 & 0.909 & 1.063 \\
\addlinespace

 & 20\% & 17.945 & 0.919 & 1.006 \\
LHM & 40\% & 17.709 & 0.916 & 1.026 \\
 & 60\% & 17.551 & 0.915 & 1.037 \\
\addlinespace

 & 20\% & \textbf{18.428} & \textbf{0.923} & \textbf{0.947} \\
LORM & 40\% & \textbf{18.199} & \textbf{0.921} & \textbf{0.965} \\
 & 60\% & \textbf{18.116} & \textbf{0.919} & \textbf{0.972} \\
\bottomrule
\end{tabular}
\vspace{-20pt}
\end{table}

\noindent \textbf{Occlusion-Robust Human Reconstruction.}
We benchmark against SOTA methods (LHM~\cite{qiu2025lhm}, PSHuman~\cite{li2025pshuman}, SyncHuman~\cite{chen2025synchuman}, IDOL~\cite{zhuang2025idol}) on naturally occluded in-the-wild images in Figure~\ref{fig:exp_fig2}. 

Lacking specific design for occlusions, these baselines often yield incomplete geometries plagued by visible transparency artifacts. 
In contrast, LORM recovers complete and structurally sound shapes, while the 3D refinement with \diffName{} further enhances local fidelity and texture sharpness. 
This confirms our pipeline's ability to ensure complete geometry and sharp textures despite severe input incompleteness.

Quantitatively, we first simulate irregular occlusions using random masks, as reported in Table~\ref{tab:lhmtune_quant}. 
Our full pipeline achieves the best PSNR, SSIM, and LPIPS, outperforming all baselines.

Second, we assess robustness under increasing occlusion ratios (20\%--60\%) in Table~\ref{tab:occ_ratio}. While LHM and IDOL suffer severe degradation as occlusion increases, LORM maintains stable performance, confirming its strong resilience.

\begin{figure}[tb] \centering
    \includegraphics[width=0.48\textwidth]{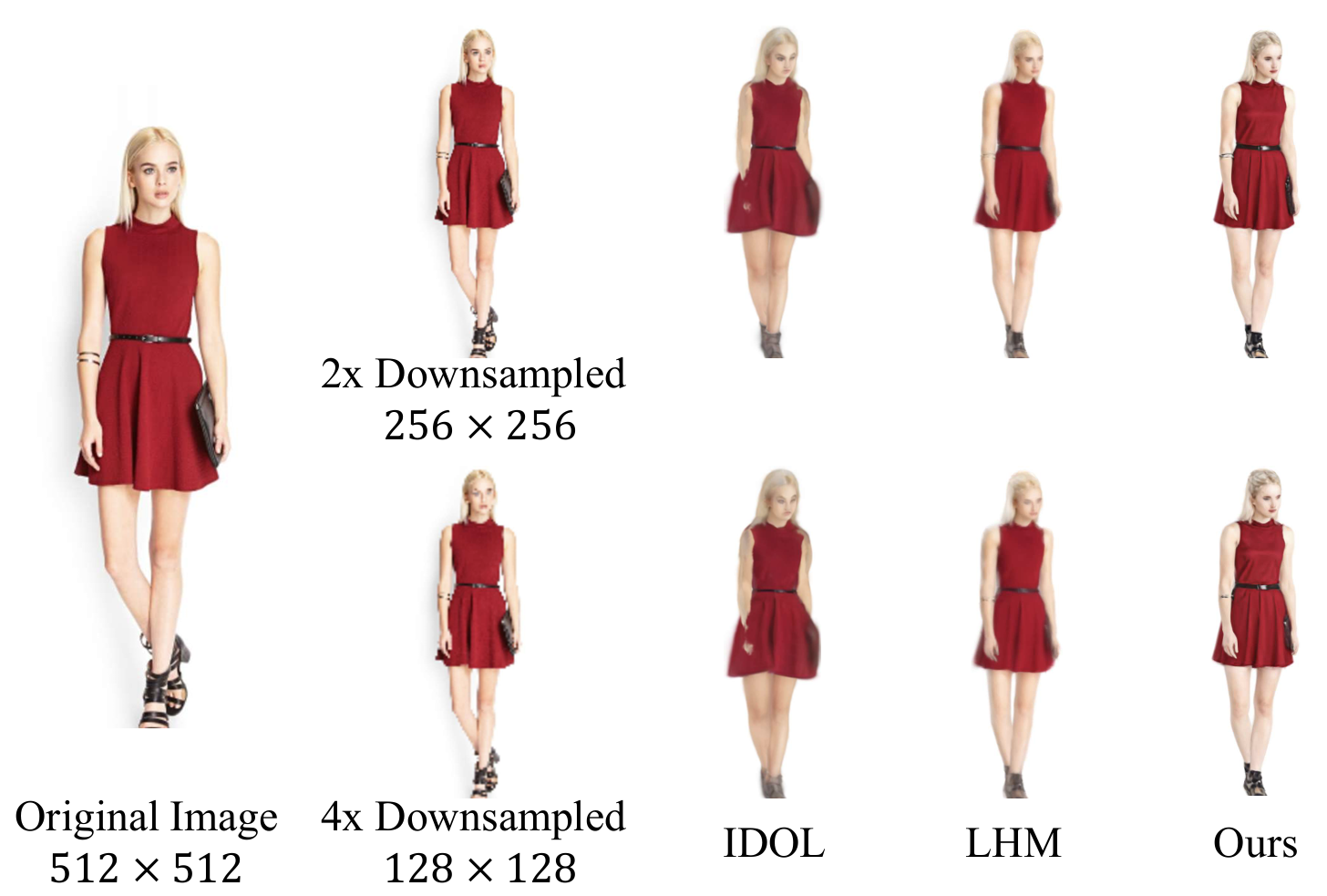}
    \vspace{-15pt}
    \caption{
Robustness to resolution degradation. The leftmost column shows the original image, followed by inputs downsampled by $2\times$ and $4\times$. Subsequent columns compare reconstructions from IDOL, LHM, and Ours. 
Even from significantly downsampled inputs, our approach consistently recovers high-fidelity geometry and sharp textures, demonstrating superior robustness to input resolution degradation compared to baselines.
    } 
    \vspace{-5pt}
    \label{fig:fig_res}
\end{figure}

\begin{table}[t]
\small
\centering
\caption{
Ablation study on the SCL strategy  and geometric conditioning inputs for the refiner. 
Results demonstrate that both SCL and the additional normal map input contribute to consistent improvements in generation quality, providing higher-fidelity supervision for the subsequent stage.
}
\label{tab:ablation_combined}
\begin{tabular}{cc ccc} 
\toprule
SCL  & Normal  & PSNR $\uparrow$ & SSIM $\uparrow$ & LPIPS $\downarrow$ \\
\midrule
$-$ & $-$ & 20.013 & 0.888 & 0.141 \\ 
$-$ & \checkmark & 20.130 & 0.892 & 0.138 \\ 
\checkmark & $-$ & 20.382 & 0.896 & 0.129 \\ 
\checkmark & \checkmark & \textbf{20.790} & \textbf{0.901} & \textbf{0.122} \\ 
\bottomrule
\end{tabular}
\vspace{-10pt}
\end{table}

\noindent \textbf{Degradation-Robust Human Reconstruction.}
To evaluate robustness against resolution degradation, we downsample the original image by $2\times$ and $4\times$ and reconstruct each using IDOL, LHM, and our full pipeline. 

As shown in Figure~\ref{fig:fig_res}, baselines suffer significantly from the resolution drop, exhibiting over-smoothed textures and boundary artifacts.
In contrast, our method leverages generative priors to recover sharper textures and coherent structures. This demonstrates that our framework effectively mitigates input degradation, enabling high-fidelity reconstruction even from severely low-resolution inputs.

\subsection{Ablation Study}
\label{sec:ablation}
We evaluate on a held-out set of 23 synthetic multi-person scenes from THuman2.1, totaling 2,622 pairs of LORM and GT mesh renderings.

\noindent \textbf{Effect of SCL Strategy.}
We evaluate the effectiveness of Self-Calibrated Learning (SCL) on the 23 test scenes. As shown in Figure~\ref{fig:SCL_vis}, without SCL, the refiner suffers from over-refinement, producing facial distortions and texture artifacts due to aggressive enhancement. In contrast, with SCL enabled, the model adaptively preserves well-recovered regions while refining incomplete areas, maintaining structural integrity.

This improvement is further validated by the quantitative results in Table~\ref{tab:ablation_combined}, where metrics evaluate the quality of the refined 2D outputs by \diffName{}. 
As shown, disabling SCL results in a noticeable performance decline across both the RGB-only and RGB+Normal configurations. 
This confirms that SCL plays a critical role in balancing refinement strength and preventing hallucinations, ensuring the generation of high-fidelity pseudo-ground truths for the subsequent 3D distillation process.

\noindent \textbf{Ablation on Geometric Conditioning.}
 We assess the importance of SMPL normal maps as 3D geometric priors. By training a variant without normal map conditioning (RGB only), we observe a clear performance drop in the refined outputs, as shown in Table~\ref{tab:ablation_combined}. This degradation, particularly in PSNR and LPIPS, confirms that explicit geometric cues are essential to prevent structural ambiguity, thereby improving pose consistency and texture alignment.
\section{Conclusion}
\label{sec:conclusion}

In this work, we present \methodName{}, a framework for single-image multi-person 3D human reconstruction. Our approach addresses occlusion and low-resolution challenges through two key components. First, LORM (Large Occluded Human Reconstruction Model) enables complete 3D geometry recovery from heavily occluded inputs via self-supervised adaptation, without requiring 3D annotations. Second, \diffName{}—a diffusion-based refiner trained with Self-Calibrated Learning (SCL)—generates high-fidelity pseudo-ground truths from coarse renderings and distills them back into the 3DGS via differentiable rendering, enhancing both local detail and global coherence.

\section{Limitations}
First, we rely on off-the-shelf estimators for pose, spatial location, and body shape; consequently, severe initialization errors may propagate to the final geometry, making hand reconstruction challenging as our pipeline is not designed to correct fundamental structural misalignments. Second, while our method recovers plausible details for occluded regions or degraded inputs, severely low resolution may lead to inconsistent identities, and these hallucinated details may not always be factually faithful to the unobserved ground truth (e.g., specific logos).

\section*{Acknowledgements}
\raggedright
This work was supported in part by the National Natural Science Foundation of China under Grant U25B2046. Fundamental and Interdisciplinary Disciplines Breakthrough Plan of the Ministry of Education of China (No. JYB2025XDXM118)
{
    \small
    \bibliographystyle{ieeenat_fullname}
    \bibliography{main}
}
\clearpage
\setcounter{page}{1}
\maketitlesupplementary
\appendix

\begin{figure}[tb] \centering
    \includegraphics[width=0.48\textwidth]{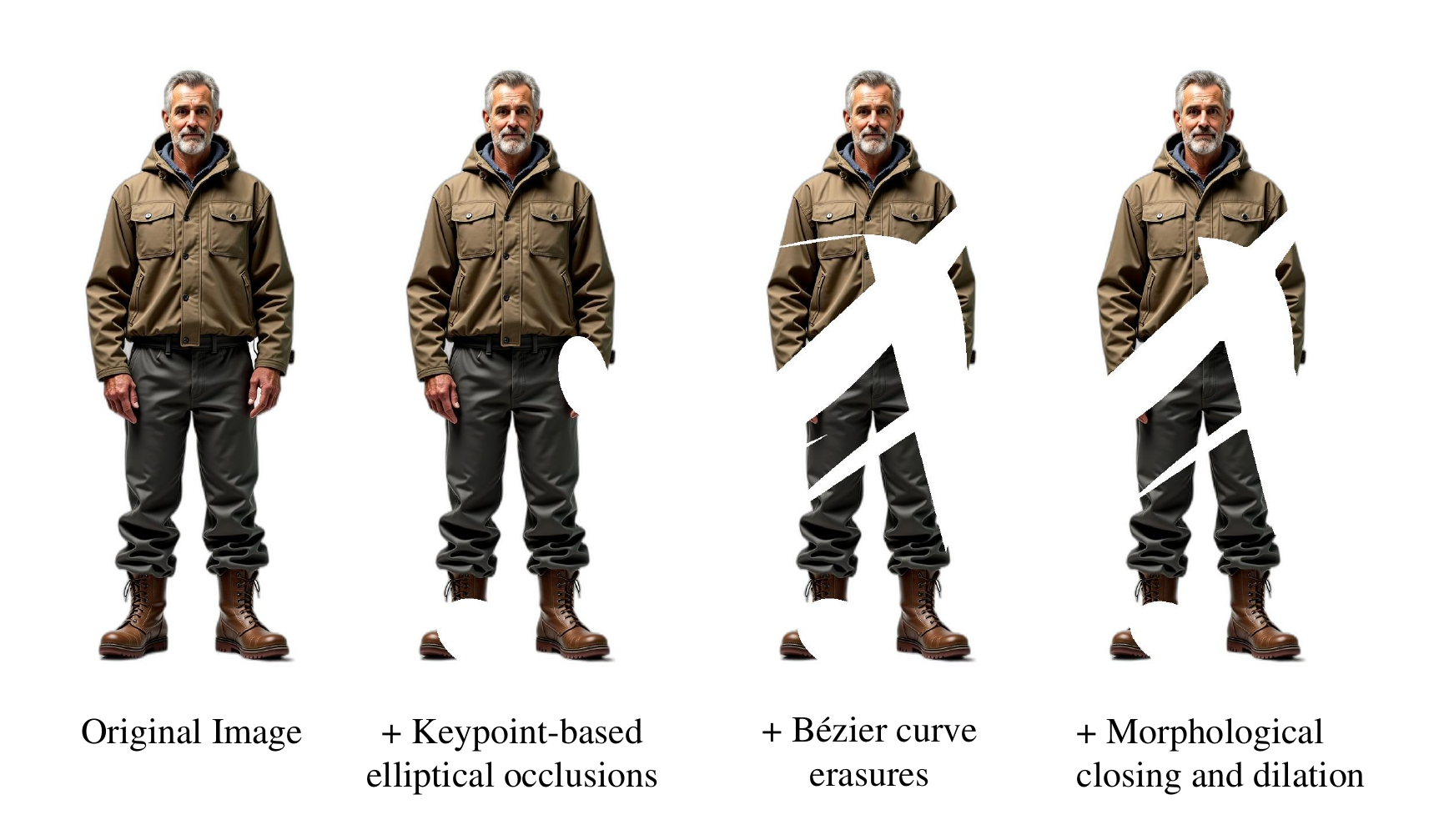}
    \vspace{-10pt}
    \caption{
    Visualization of the multi-stage occlusion mask generation process. 
    From left to right: original image, with keypoint-based elliptical occlusions applied, further corrupted by irregular Bézier-based erasures, and finally smoothed via morphological operations. 
    }
    \vspace{-10pt}
    \label{fig:vis_mask}
\end{figure}

\begin{figure}[tb] \centering
    \includegraphics[width=0.48\textwidth]{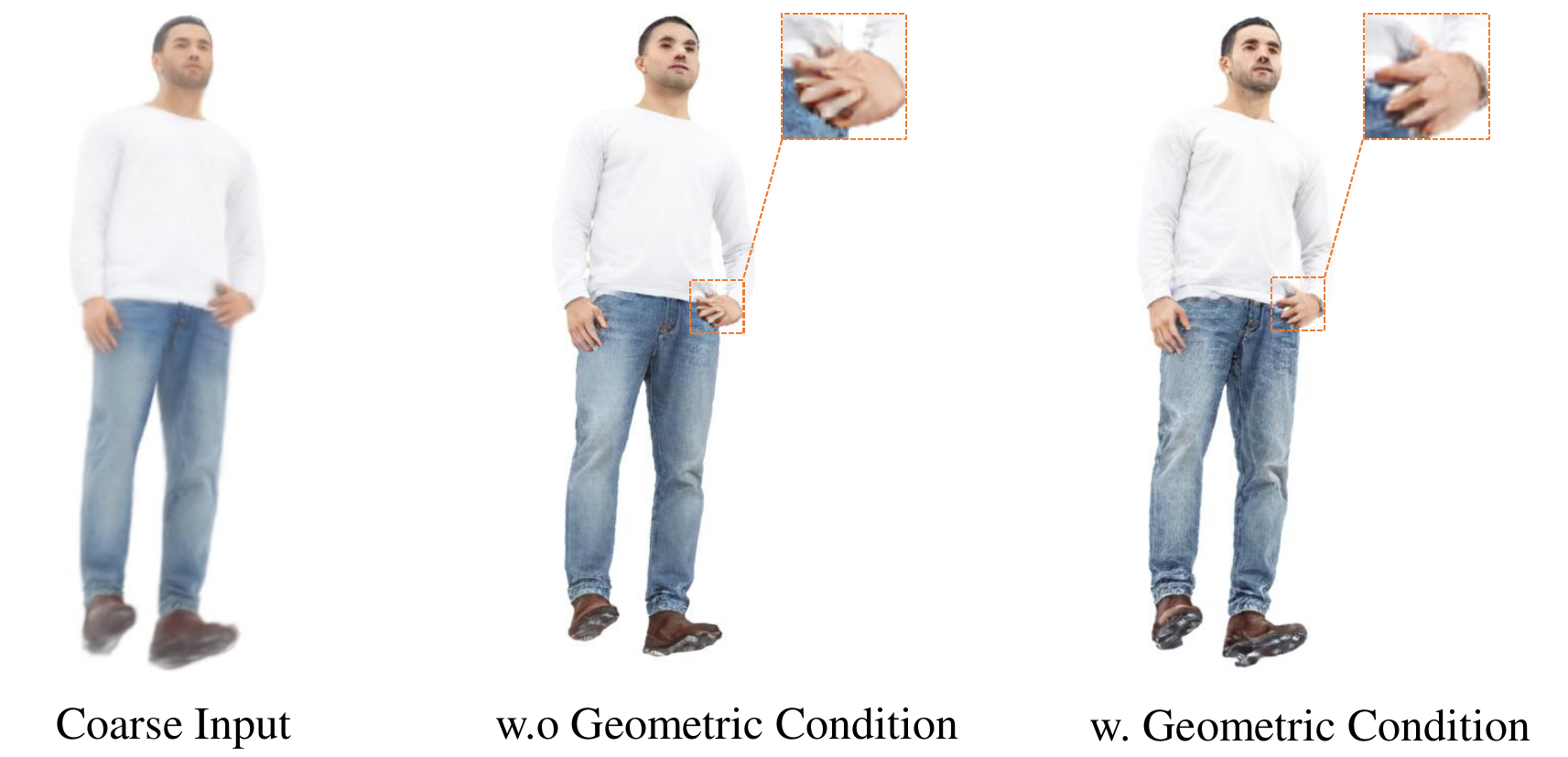}
    \vspace{-10pt}
    \caption{
    Ablation on geometric conditioning using SMPL normal maps. 
    Left: Coarse input. 
    Middle: Result without normal map conditioning — suffers from hand collapse and facial distortion. 
    Right: Result with normal map — preserves correct hand structure and facial geometry. 
    Zoom-ins show significant improvement in local detail fidelity and structural coherence.
    } 
    \label{fig:normal_ablation}
\end{figure}

\section{Occlusion Simulation for Self-Supervised Training}
To ensure robust generalization under real-world occlusions, we design a structured yet diverse masking strategy to synthesize $I_{\text{occ}}$ from the full image $I_{\text{full}}$. The occlusion process combines three complementary components: anatomically plausible local erasures, global irregular masks, and structural line cuts—ensuring both realism and diversity in training data.

First, we detect human body keypoints using MediaPipe Pose~\cite{lugaresi2019mediapipe}, excluding head landmarks to focus on torso and limb regions. For each selected keypoint $\mathbf{p}_k = (x_k, y_k)$, we apply an elliptical occlusion:
\begin{equation}
\mathcal{M}_{\text{ellipse}}^{(k)} = \left\{ (x,y) \in \mathbb{R}^2 \,\middle|\, \frac{(x - x_k)^2}{a_x^2} + \frac{(y - y_k)^2}{a_y^2} \leq 1 \right\},
\end{equation}
where semi-axes $a_x, a_y \sim \mathcal{U}(30, 100)$ and rotation angle $\theta \sim \mathcal{U}(0^\circ, 360^\circ)$ are sampled randomly. Up to $K=5$ such ellipses are applied, with number and location randomized per instance.

Second, we generate large-scale irregular masks using Bézier curves. A quadratic Bézier path is defined by three control points $\mathbf{C}_0, \mathbf{C}_1, \mathbf{C}_2 \in [0,H]\times[0,W]$, uniformly sampled across the image. The curve is thickened into a filled region via polygon rasterization, forming a continuous occluding band:
\begin{equation}
\mathcal{M}_{\text{bezier}} = \bigcup_{i=1}^{N_b} \text{FillPoly}(\text{Bezier}(\mathbf{C}_0^{(i)}, \mathbf{C}_1^{(i)}, \mathbf{C}_2^{(i)}, t),\, t\in[0,1]),
\end{equation}
where $N_b \sim \mathcal{U}\{0,5\}$ controls the number of such global erasures.

Third, with probability $p=0.5$, we apply a random straight-line cut that partitions the image into two half-planes. Given two random points $(x_1,y_1)$, $(x_2,y_2)$, the dividing line is:
\begin{equation}
L(x,y) = (y - y_1)(x_2 - x_1) - (x - x_1)(y_2 - y_1) > 0.
\end{equation}
We select one side for occlusion, ensuring the masked area does not exceed 70\% of the total to preserve sufficient visible content.

The final binary mask $\mathcal{M}_{\text{final}}$ is obtained by combining all components:
\begin{equation}
\mathcal{M}_{\text{final}} = \mathcal{M}_{\text{ellipse}} \cup \mathcal{M}_{\text{bezier}} \cup \mathcal{M}_{\text{line}},
\end{equation}
followed by morphological closing and dilation (kernel size $5\times5$, iterations=3) to smooth boundaries and simulate realistic cloth or object overlap.

As illustrated in Figure~\ref{fig:vis_mask}, this multi-step process progressively constructs complex occlusions that mimic real-world scenarios—such as one person being partially blocked by another or obscured by foreground objects.

This multi-level occlusion scheme ensures that the student model (LORM) learns to recover missing geometry under diverse and challenging conditions, while the frozen teacher provides consistent pseudo-ground truths for stable self-distillation. During training, we render the 3DGS representation from 24 horizontally distributed camera viewpoints uniformly placed on a full 360-degree orbit around the subject, ensuring comprehensive geometric coverage for robust self-supervised learning.

\section{Effect of Geometric Conditioning}
To evaluate the impact of geometric priors in diffusion-based refinement, we ablate the use of SMPL normal maps as conditioning input. As shown in Figure~\ref{fig:normal_ablation}, without geometric guidance (middle), the refiner often produces structural distortions—such as collapsed hands or misaligned limbs—due to ambiguous depth and pose cues in coarse renderings. In contrast, when conditioned on the SMPL normal map (right), our method preserves anatomically plausible hand and facial structures, demonstrating improved 3D consistency.

The zoom-in views highlight that geometric conditioning helps recover fine-scale details  that are otherwise hallucinated or distorted. This confirms that the normal map provides effective 3D-aware inductive bias, guiding the diffusion process to respect human body geometry during refinement.

\begin{figure*}[tb] \centering
    \includegraphics[width=\textwidth]{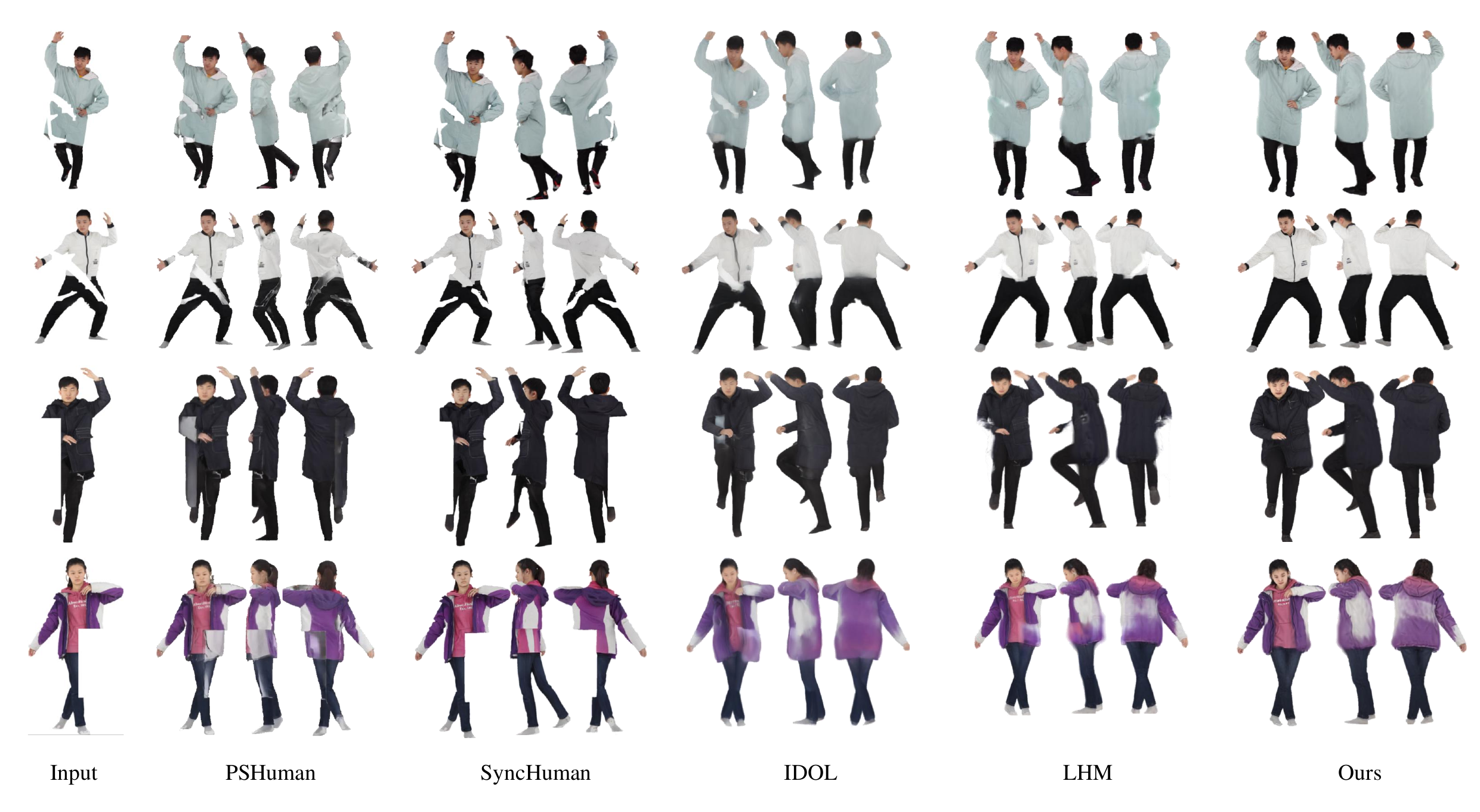}
    \vspace{-5pt}
    \caption{
    Qualitative comparison on occluded human reconstruction (THuman2.1). 
    Input images are occluded via Bézier curves or rectangular masks. 
    } \label{fig:exp_fig3}
    \vspace{-5pt}
\end{figure*}

\begin{figure*}[tb] \centering
    \includegraphics[width=\textwidth]{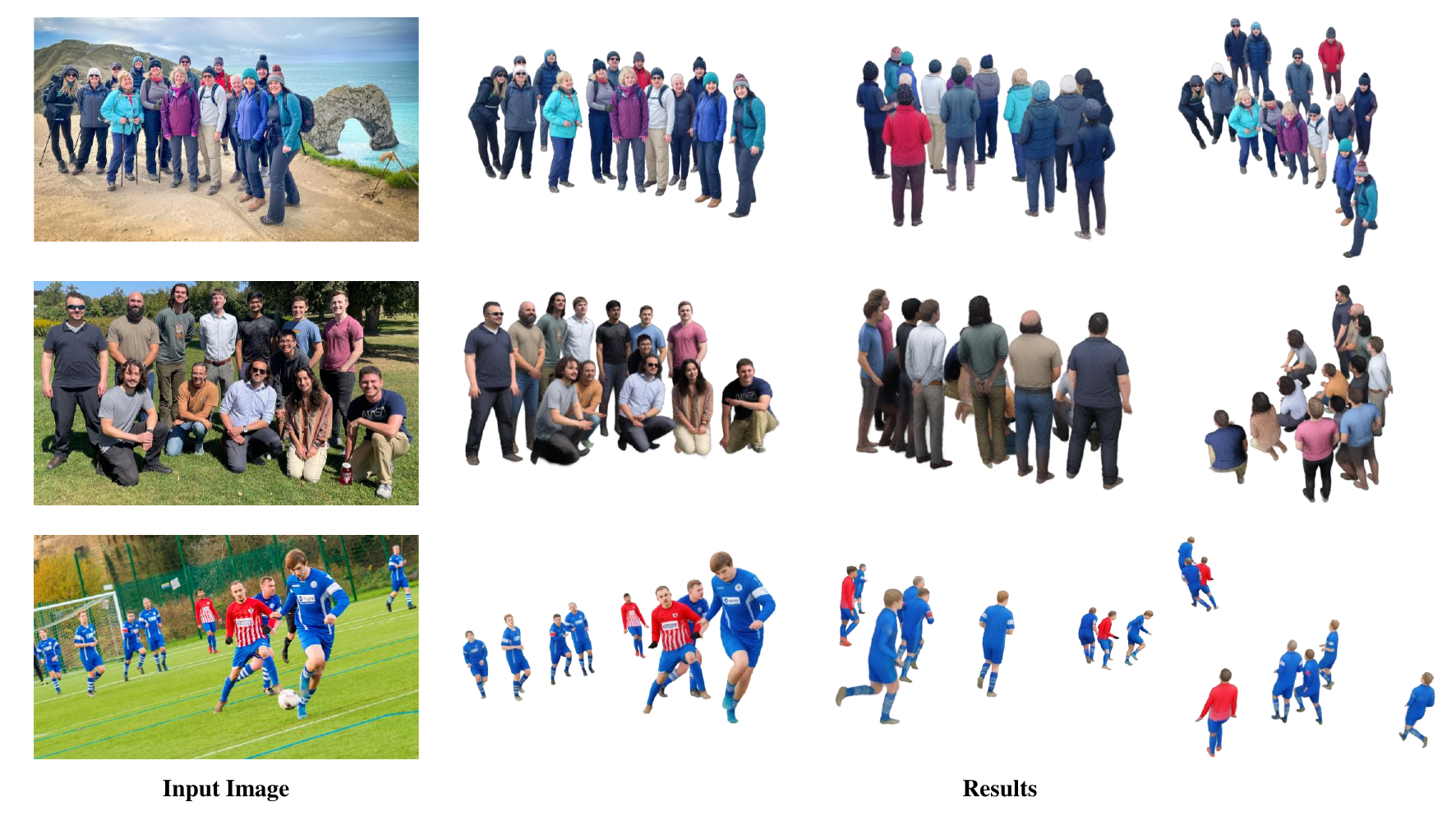}
    \vspace{-5pt}
    \caption{
    More reconstruction results on diverse in-the-wild multi-person images.
    } \label{fig:exp_fig4}
    \vspace{-5pt}
\end{figure*}

\section{Limitations of 2D Inpainting as Preprocessing}
A natural alternative to our self-supervised adaptation framework is to first apply 2D image inpainting to occluded person crops and then feed the completed image into a pre-trained 3D generator such as LHM~\cite{qiu2025lhm}. However, we find this strategy ineffective due to the inherent limitations of 2D inpainting under real-world occlusions.

\begin{figure} \centering
    \includegraphics[width=0.48\textwidth]{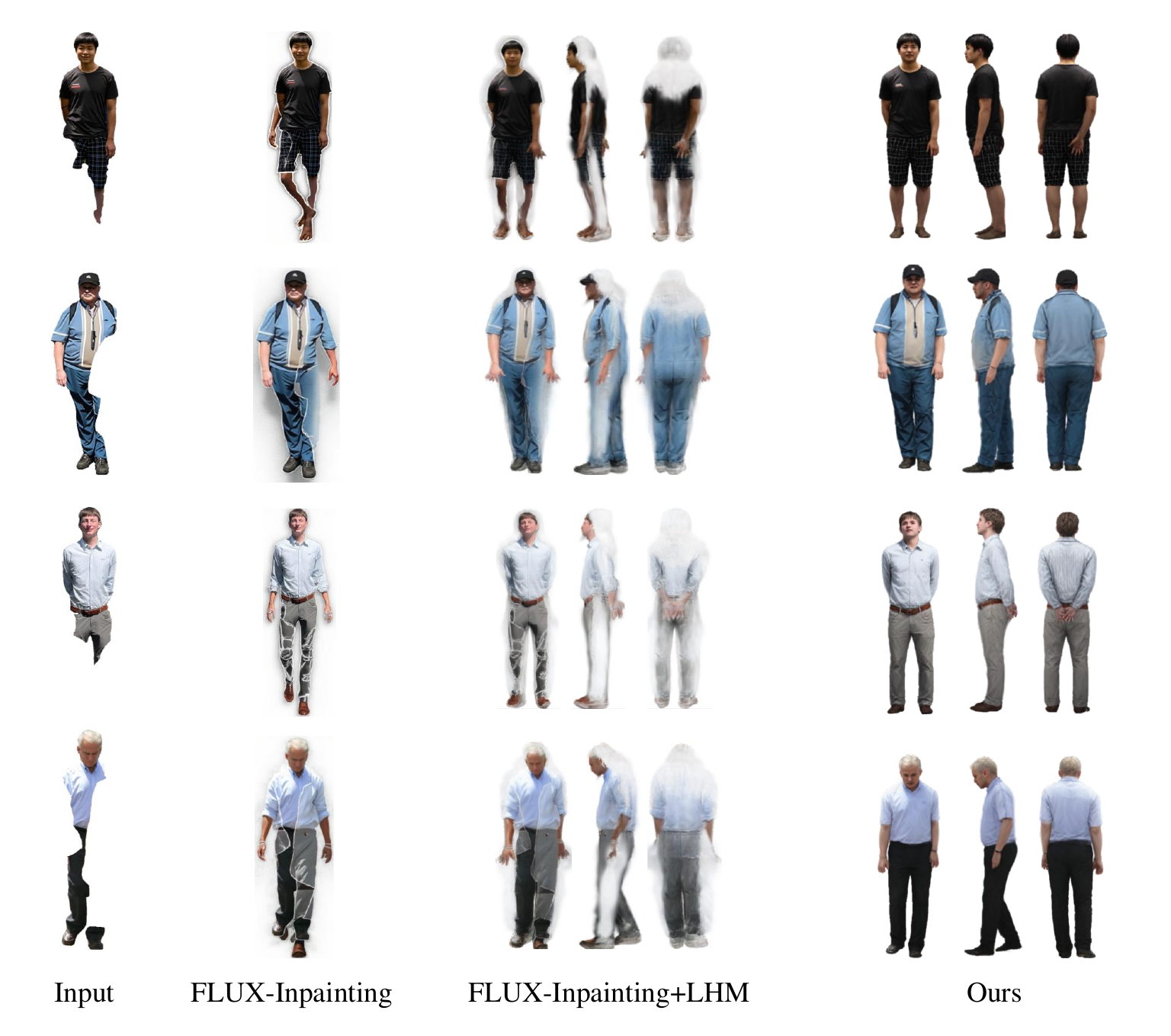}
    \vspace{-10pt}
    \caption{
    Comparison with the 2D inpainting + LHM pipeline. 
    } 
    \vspace{-10pt}
    \label{fig:exp_fig6}
\end{figure}

As shown in Figure~\ref{fig:exp_fig6}, while modern diffusion-based inpainters~\cite{flux-inpaint} can generate visually plausible textures locally, they often produce boundary artifacts, inconsistent lighting, and semantically implausible content—such as incorrect limb structures or mismatched clothing patterns. These errors are not only visually jarring but also mislead the downstream 3D reconstruction model.

When such inpainted images are passed to LHM, the generated 3DGS exhibits amplified artifacts, including distorted geometry and texture inconsistencies across views. This confirms that 2D inpainting lacks 3D-awareness and cannot reliably recover structurally coherent human shapes under arbitrary occlusion patterns.

\section{More Results}

We present additional qualitative results to further demonstrate the effectiveness and generalization of our method.

Figure~\ref{fig:exp_fig3} compares our method against recent state-of-the-art approaches—including LHM~\cite{qiu2025lhm}, PSHuman~\cite{li2025pshuman}, SyncHuman~\cite{chen2025synchuman}, and IDOL~\cite{zhuang2025idol}—on occluded reconstructions from THuman2.1~\cite{zheng2019deephuman}. The occlusions are synthesized using either Bézier curves or random rectangular masks.

Figure~\ref{fig:exp_fig4} shows reconstructions on a variety of challenging in-the-wild images containing multiple individuals.

\section{Multi-Person Human Mesh Recovery}
\label{sec:supp_pose_shape}

This section provides background on parametric human modeling and multi-person 3D human recovery from a single image, which forms the foundation of our pipeline.

\paragraph{Human Parametric Model}
\label{sec:preliminary-human-model}
The SMPL~\cite{loper2023smpl} and SMPL-X parametric model~\cite{loper2023smpl, pavlakos2019expressive} are widely used for representing 3D human body shape and pose. It represents a deformable human mesh as a function of low-dimensional parameters:
\begin{equation}
\mathcal{M}(\boldsymbol{\beta}, \boldsymbol{\theta}) = \mathcal{W} \left( 
\mathcal{V}(\boldsymbol{\beta}, \boldsymbol{\theta}), 
\mathcal{J}(\boldsymbol{\beta}), 
\mathcal{T}_p(\boldsymbol{\theta}), 
\mathcal{W} 
\right),
\end{equation}
where $\boldsymbol{\beta} \in \mathbb{R}^{10}$ denotes the shape parameters (learned from PCA over 3D scans), and $\boldsymbol{\theta} \in \mathbb{R}^{23 \times 3}$ represents the joint rotation angles in axis-angle form. The model applies linear blend skinning ($\mathcal{W}$) to a template mesh $\mathcal{V}$, with pose-dependent blend shapes and joint regressor $\mathcal{J}(\boldsymbol{\beta})$ to generate realistic articulations.

\paragraph{Multi-Person Human Mesh Recovery from a Single Image}
\label{sec:multi_smpl_estimation}
To recover multiple humans from a single image, recent methods~\cite{baradel2024multi, wang2025prompthmr} estimate per-person SMPL-X parameters along with their camera-space positions. Given an input image $I$, the goal is to infer a set of parameters for each detected individual $i = 1,\dots,N$:
\begin{equation}
\left\{ \boldsymbol{\theta}_i, \boldsymbol{\beta}_i, \mathbf{t}_i \right\}_{i=1}^N = f_{\text{pose}}(I),
\end{equation}
where $\boldsymbol{\theta}_i$ and $\boldsymbol{\beta}_i$ are the pose and shape parameters of person $i$, and $\mathbf{t}_i \in \mathbb{R}^3$ is the root translation (camera-relative position).

In our pipeline, we adopt PromptHMR~\cite{wang2025prompthmr} to extract these parameters, which provides robust multi-person pose and global positioning, serving as input for subsequent processing stages.

\section{Implementation Details}

\paragraph{LORM Training.}

For efficient fine-tuning, we update the transformer block using LoRA with rank 32, while other inherited pre-trained components remain frozen. The model is trained for approximately 2,000 iterations using the AdamW~\cite{kingma2014adam} optimizer with a learning rate of  $1 \times 10^{-4}$.

\paragraph{CrowdRefiner Training.}

We build \diffName{} based on SD-Turbo~\cite{sauer2024adversarial}, employing a single-step diffusion process with noise level $\tau=200$ (maximum diffusion time $\tau=1000$), enabling efficient training and inference. We apply LoRA to the VAE decoder with rank 4, while keeping the encoder frozen. The model is trained with the composite loss $\mathcal{L}_{\text{diffuse}}$ for approximately 10,000 steps with a batch size of 2. We apply the SCL strategy by replacing the input with $R_{\text{gt}}$ at probability $\rho = 0.2$ during training. The model is optimized using AdamW~\cite{kingma2014adam} with a learning rate of $2 \times 10^{-5}$.

\paragraph{Evaluation Setup on THuman2.1}
For quantitative evaluation on occluded human reconstruction, we select 500 samples from THuman2.1~\cite{zheng2019deephuman}. Ground truth renderings are generated by rendering the high-fidelity meshes from 24 uniformly distributed viewpoints along a full 360-degree horizontal orbit. These views serve as reference images for computing PSNR, SSIM, and LPIPS to assess reconstruction quality.

\section{Evaluation Metrics}
\label{sec:supp_metrics}

We evaluate the performance of \diffName{} and compare our method with baselines on occluded human reconstruction using three widely adopted image quality metrics: PSNR, SSIM, and LPIPS. These provide complementary insights into reconstruction fidelity—from pixel accuracy to perceptual similarity.

\paragraph{PSNR} Peak Signal-to-Noise Ratio (PSNR) measures the pixel-wise similarity between the predicted rendering $I_{\text{pred}}$ and the ground truth $I_{\text{gt}}$. It is computed based on the mean squared error (MSE):
\begin{equation}
\text{MSE} = \frac{1}{N} \sum_{p} (I_{\text{pred}}[p] - I_{\text{gt}}[p])^2,
\end{equation}
where $p$ indexes all pixels. The PSNR is then:
\begin{equation}
\text{PSNR} = 10 \cdot \log_{10}\left( \frac{\text{MAX}^2}{\text{MSE}} \right),
\end{equation}
with $\text{MAX}$ as the maximum pixel value (e.g., 255). Higher values indicate better photometric accuracy.

\paragraph{SSIM} Structural Similarity (SSIM) evaluates how well structural patterns—such as edges and textures—are preserved. Instead of global statistics, it computes similarity in local windows using luminance, contrast, and structure comparisons:
\begin{equation}
\text{SSIM}(I_{\text{pred}}, I_{\text{gt}}) = \frac{(2\mu_p\mu_g + C_1)(2\sigma_{pg} + C_2)}{(\mu_p^2 + \mu_g^2 + C_1)(\sigma_p^2 + \sigma_g^2 + C_2)},
\end{equation}
where $\mu$, $\sigma$ denote local means and standard deviations, and $\sigma_{pg}$ is the cross-covariance. Constants $C_1, C_2$ prevent division by zero. The final score is averaged over all patches. Values closer to 1 indicate stronger structural consistency.

\paragraph{LPIPS} Learned Perceptual Image Patch Similarity (LPIPS) assesses perceptual quality by comparing deep features from a pre-trained network. Rather than raw pixels, it operates in feature space:
\begin{equation}
\text{LPIPS}(I_{\text{pred}}, I_{\text{gt}}) = \sum_l w_l \cdot \| \phi_l(I_{\text{pred}}) - \phi_l(I_{\text{gt}}) \|_2^2,
\end{equation}
where $\phi_l$ extracts features from layer $l$ of a VGG-16~\cite{simonyan2014very} backbone, and $w_l$ are learned weights that emphasize semantically meaningful layers. Lower LPIPS values correspond to higher perceptual similarity, aligning closely with human judgment.

Together, these metrics allow us to assess reconstructions from multiple perspectives: PSNR for pixel-level accuracy, SSIM for structural coherence, and LPIPS for high-level visual realism.

\section{Source of In-the-Wild Images}
\label{sec:source_wild_images}

The in-the-wild multi-person images used in our qualitative evaluation are collected from publicly available sources on the web, including search engines such as Google Images~\cite{google-images}. These images are selected solely for visual demonstration and are not used in quantitative benchmarking. We ensure that all displayed examples fall under fair use for academic illustration, with no commercial intent. For reproducibility, we do not claim ownership of these images and encourage readers to refer to the original sources via the search engine.


\end{document}